\documentclass{article}
\usepackage{spconf,amsmath,graphicx}
\usepackage{indentfirst}
\usepackage[pagebackref=true,breaklinks=true,letterpaper=true,colorlinks,bookmarks=false]{hyperref}
\usepackage{mathtools}

\usepackage{amssymb}
\usepackage{multirow}
\usepackage{siunitx}
\usepackage{algorithmic}

\usepackage{color, colortbl}
\definecolor{LightCyan}{rgb}{0.88,1,1}
\definecolor{LightRed}{rgb}{1,0.88,0.95}
\usepackage[boxed]{algorithm2e}
\SetAlCapSkip{1em}
\usepackage{comment}
\usepackage{amsmath}
\usepackage{bm}

\SetKw{Continue}{continue}

\title{Context-Aware Data Augmentation for LiDAR 3D Object Detection}
\name{Xuzhong Hu\sthanks{E-mail: huxuzhong@hust.edu.cn}, Zaipeng Duan, and Jie Ma\sthanks{Corresponding Author E-mail: majie@hust.edu.cn}}
\address{}
\begin{document}
	\ninept
	\maketitle
	\begin{abstract}
		For 3D object detection, labeling lidar point cloud is difficult, so data augmentation is an important module to make full use of precious annotated data. As a widely used data augmentation method, GT-sample effectively improves detection performance by inserting groundtruths into the lidar frame during training. However, these samples are often placed in unreasonable areas, which misleads model to learn the wrong context information between targets and backgrounds. To address this problem, in this paper, we propose a context-aware data augmentation method (CA-aug) , which ensures the reasonable placement of inserted objects by calculating the "Validspace" of the lidar point cloud. CA-aug is lightweight and compatible with other augmentation methods. Compared with the GT-sample and the similar method in Lidar-aug(SOTA), it brings higher accuracy to the existing detectors. We also present an in-depth study of augmentation methods for the range-view-based(RV-based) models and find that CA-aug can fully exploit the potential of RV-based networks. The experiment on KITTI val split shows that CA-aug can improve the mAP of the test model by 8\%.
	\end{abstract}

	\begin{keywords}
		3D object detection, lidar, augmentation
	\end{keywords}

	\section{Introduction}
	\label{sec:intro}
	In the automatic driving system, 3D detection is the basis of object tracking, path planning and other functions. Due to the ability to directly obtain accurate spatial geometric information and the continuous reduction of costs, vehicle-mounted lidar plays an important role in the scene perception of autonomous driving. Therefore, 3D object detection in lidar point cloud has attracted the attention of many researchers. In recent years, thanks to autonomous driving datasets such as KITTI and Waymo,3D detection models have been developed rapidly. Their accuracy and reliability have been continuously improved. However, compared with images, labeling 3D training data is more costly and time-consuming, due to the sparsity of point clouds and the higher degree of freedom of bounding boxes. However, neural networks require a large amount of annotated data to ensure accurate inference results.
	
	One solution to this problem is to generate more training samples by data simulators such as Carla\cite{dosovitskiyCARLAOpenUrban2017} and Airsim\cite{shahAirSimHighFidelityVisual2017}.To deal with the large sim-to-real domain gap, some researchers combine scan data from real-world scenarios with CAD models. For example, \cite{fangAugmentedLiDARSimulator2020} built high-resolution static maps using the professional 3D scanner Riegl VMX-1HA, from which low-resolution scene point cloud data can be obtained with CAD target models inserting into it. In Lidar-smi\cite{manivasagamLiDARsimRealisticLiDAR2020}, the authors built 3D maps and object models via multi-frame registration and surface reconstruction with no need for expensive laser scanners.
	
	\begin{figure}[t!]
		\centering
		\begin{tabular}{cc}
			\includegraphics[width=0.45\columnwidth]{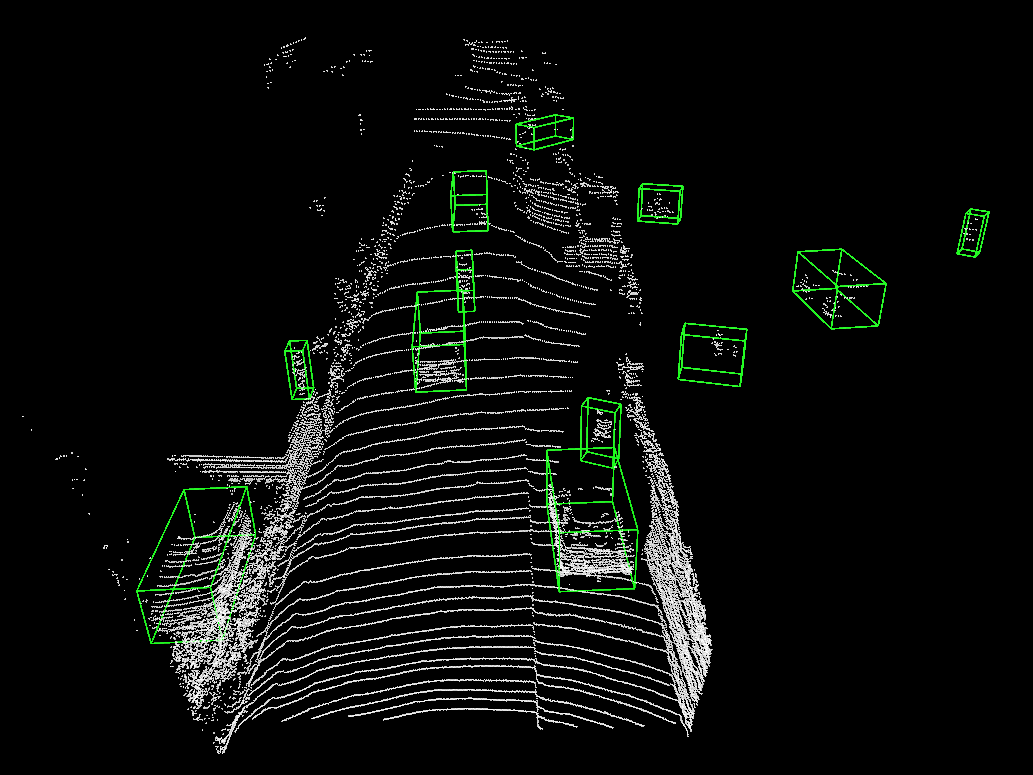}
			&
			\includegraphics[width=0.45\columnwidth]{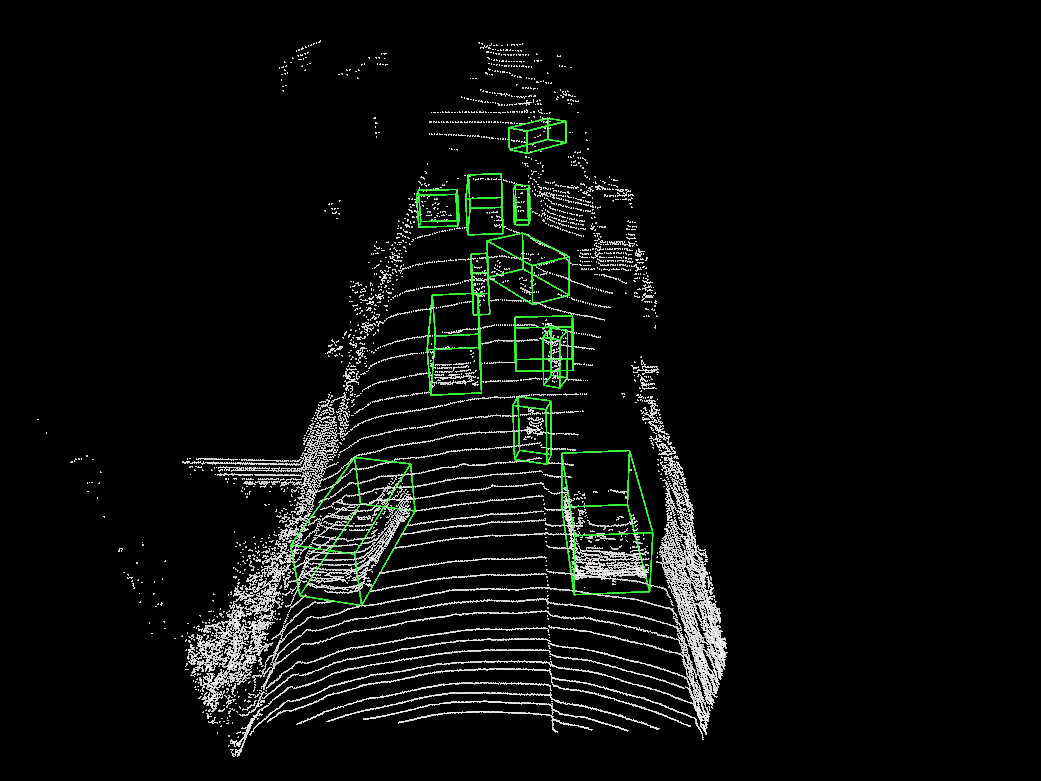}\\
			(a) & (b)
		\end{tabular}
		\begin{tabular}{c}
			\includegraphics[width=0.9\columnwidth]{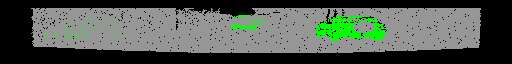}\\
			(c) \\
			\includegraphics[width=0.9\columnwidth]{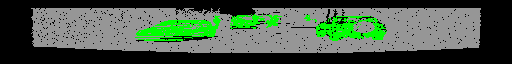}\\
			(d) \\
		\end{tabular}
		\caption{Comparsion of two augmentations. (a) In GT-sample, objects are inserted into the scene at their original positions. (b) CA-aug can ensure the reasonable placement of inserted objects. (c) the range image of the point cloud scene augmented by GT-Sample. (d) After CA-aug , more object points(green) are retained.}
		\label{fig:show}
	\end{figure}

	\begin{figure*}[ht!]
		\centering
		\begin{tabular}{ccc}
			\includegraphics[width=1.8\columnwidth]{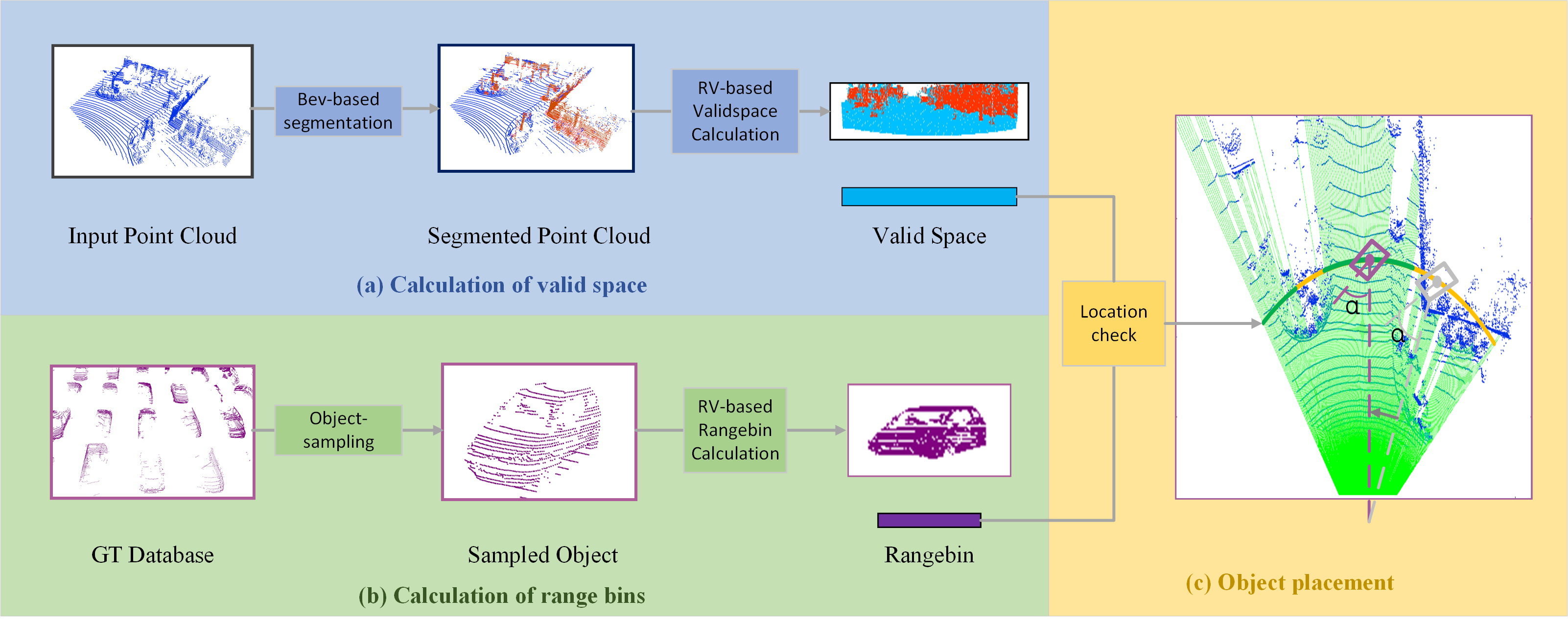}
		\end{tabular}
		\caption{The framework of CA-Aug. (a) The input point cloud is split into ground points (blue) and obstacle points (red) and they are projected into a range image from which "Validspace" can be obtained; (b) Randomly sample a target from the GT database and the "Rangebin" can be calculated from its range image; (c) Feasible rotation angles calculated with "Validspace" and "Rangebin" as indicated by the green part of the arc and the purple box indicates the selected pose of the inserted target while the gray one is the original position.}
		\label{fig:flowchart}
	\end{figure*}

	Unlike data simulation, data augmentation does not require much time to generate synthetic training samples in advance. It can be integrated into the training process and effectively avoid overfitting. Similar with 2D object detection, basic data augmentations\cite{hahnerQuantifyingDataAugmentation2020}, such as global rotation, mirroring, and translation, is widely used in point cloud. To increase the flexibility of data augmentation, some researchers\cite{Cheng2020Improving3O,liPointAugmentAutoAugmentationFramework2020} use optimistic strateties to automate relative hyperparameters. Adversarial learning can also be employed to adjust the difficulty of training samples during training to improve model robustness\cite{maPointDropImprovingObject2021}. In GT-sample\cite{yanSECONDSparselyEmbedded2018}, the author created a database for the information of all groundtruths. During training, some objects were randomly selected from it and inserted into the point cloud scene. Due to its ability to rectify the extreme data imbalance between the positive and negative examples and significantly improve the performance of detectors, GT-sample is widely used in various models. After that, a series of new methods\cite{choiPartAwareDataAugmentation2021,choiPartAwareDataAugmentation2021,zhangGLENetBoosting3D2022,zhengSESSDSelfEnsemblingSingleStage2021} were proposed. Most of them improved the generalization of the network by changing the distribution of surface points of groundtruths, which was equivalent to extend the groundtruth database. 
	 
	However, in GT-sample, the objects are inserted into the lidar frame with their original positions, which means they are likely to appear in unreasonable areas, such as in a wall or behind a building. It has two bad effects on learning: (1) Because the poses of inserted objects have nothing to do with the current scene, the semantic information between them loses. (2) Compared with vehicles, pedestrians and cyclists surface points are sparser and more irregular in shape. When they appear in unreasonable areas, the network is difficult to distinguish them from noises, which will mislead the neural network to learn error information.
	
	In recent years, detection methods\cite{Chai2021ToTP,meyerLaserNetEfficientProbabilistic2019} based on range view(RV) have attracted the attention of many researchers for easy deployment and fast inference speed. However, RV-based models require maintaining the 2.5D structure of the lidar point cloud. In the process of spherical projection, the occluded points need to be discarded. Besides above two disadvantages, excessive occlusion will destroy the surface structure of inserted objects, thus bringing noises to the network during training. We divide existing augmentation strategies into three categories according to \cite{huWhatYouSee2020}. (1) The point cloud augmented by GT-sample is directly projected into the range image\cite{guCVFNetRealtime3D2022,liangRangeRCNNFastAccurate2021}. (2) Remove the inserted objects which loss too many points due to occlusion\cite{fanRangeDetDefenseRange2021}. (3) Keep the objects point as many as possible by deleting background points in the same pixel of the range image.
	
	As shown in Figure \ref{fig:show}, to solve the above problems, we propose a context-aware 3D data augmentation CA-aug. By dividing the lidar data into ground points and obstacle points, it is easy to calculate the "Validspace" where targets should place. Note that we are not the first to attempt to do so. Lidar-aug\cite{fangLiDARAugGeneralRenderingbased2021} has proposed the similar idea by constructing  "Validmap" which consists of many pillars. During training, these pillars are uniformly sampled and the positions of objects and CAD models are randomly selected in them. However, we further consider the different surface point distribution of objects. Experiments show that our method is superior to GT-sample and  the similar idea in Lidar-aug in detection accuracy improvement. We also demonstrate that CA-aug can control the target occlusion and significantly improve the performance of RV-based models.
	
	In summary, the main contributions in this work are as follows:
	\begin{itemize}
		\item We propose a context-aware object augmentation method that solves the problem of unreasonable placement of inserted samples in GT-sample and significantly improves the accuracy of different models, especially for the detection of pedestrians and cyclists.
		
		\item We study the augmentation of RV-based detectors that few people have noticed. Compared to other methods, CA-aug can fully exploit the potential of the RV-based model and achieve 8\% mAP improvement on the detection of moderate targets in the KITTI val set.
		
		\item Our algorithm is lightweight and compatible with existing augmentation methods.
	\end{itemize}

	\section{METHOD}
	\label{sec:method}
	\vspace{-0.3cm}
	\subsection{Overview}
	\label{overview}
	
\begin{figure}[t!]
		\centering
		\begin{tabular}{cc}
			\includegraphics[width=0.4\columnwidth]{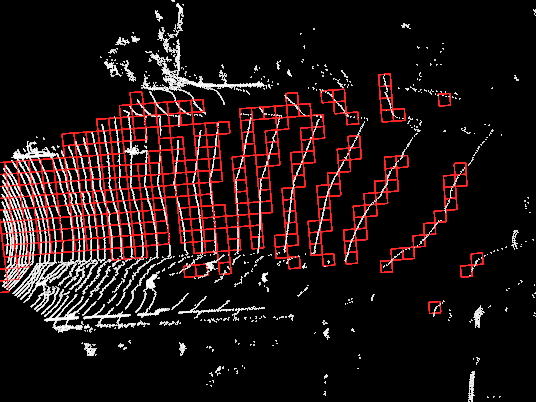}
			&
			\includegraphics[width=0.4\columnwidth]{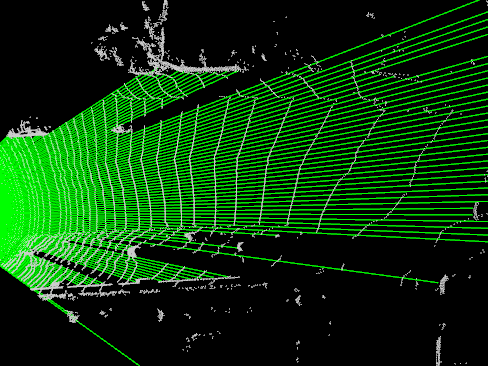}\\
			(a) & (b)
		\end{tabular}
		\caption{(a) Red grids represent the "Validmap" of the scene. (b) The green rays represent the "Validspace" of the scene.}
		\label{fig:freespace}
	\end{figure}

	It is non-trivial to find reasonable places for objects in the scene point cloud because background points in the detection dataset are not labeled. However, we can simplify this problem to placing targets where the scan line can reach. In \cite{huWhatYouSee2020}, the authors use a 3D voxelized visibility map to record the area where laser beams pass through, but it requires much computation and memory cost. As shown in Figure \ref{fig:flowchart}(a), we propose a simple approach. We assume that the raw point cloud is composed of ground points and obstacle points and project them into the range image. For each column, the part from bottom to the nearest obstacle point is where the target points are allowed to appear. All the parts constitute "Validspace" which can be simply represented by a vector. As shown in Figure \ref{fig:freespace}(b), it can also cover the pointless area between two rows of scan lines.
	
	If the location of each inserted target is randomly selected from "Validspace", the relationship between the density of the surface points and the distance to the lidar will be broken. Therefore, like \cite{sebekReal3DAugPointCloud2022}, we keep the original range and rotate the target around the vertical z-axis to the right position. The task is to determine each target's rotation angle. We also project the target point to the range image and use a vector "Rangebin" to describe its points pattern, which does not change when rotating around the z-axis. Both "Validspace" and "Rangebin" can be calculated before training. As shown in Figure \ref{fig:flowchart}(c), the location of each object can be obtained from them. For RV-based models, "Culling" is applied after object placement.
	\vspace{-0.2cm}
	\subsection{Calculation of Validspace and Rangebin}
	Inspired by \cite{fangLiDARAugGeneralRenderingbased2021}, We divide input point cloud into different $d\times d$ pillars. And obstacle pillars $S^o$ can be obtained as following strategy:
	\begin{equation}
		{S^o} = \{ {s_i}|\max ({z_p}) - \min ({z_p})>\sigma,p{\kern 1pt} {\kern 1pt} {\rm{in}}{\kern 1pt} {\kern 1pt} {s_i}\}
	\end{equation} 

	Where $z_p$ is the z coordinate of p and $s_i$ is the $i$th pillar. All points in $S^o$ are seen as obstacle points.
	
	Mechanical lidars have fixed horizontal and vertical angular resolution, so the point cloud can be projected into the range image with size of $W\times{H}$ through the following equation:
	\begin{equation}
		\left( {\begin{array}{*{20}{c}}
				u\\
				v
		\end{array}} \right) = \left( {\begin{array}{*{20}{c}}
				{1/2[1 - \arctan (y,x)/\pi ]W}\\
				{[1 - (\arcsin (z/r) - {f_{down}})/f]H}
		\end{array}} \right)
	\end{equation}

	Where $f=f_{up}-f_{down}$ is the vertical field of view, $r=\sqrt{x^2+y^2+z^2}$ is the range of point.
	
	From the range image, "Validspace" $\bm{V}$ can be obtained:
	\begin{equation}
		\bm{V}[j] = \left\{ {\begin{array}{*{20}{c}}
				{\min ({r_p}),{\kern 1pt} {\kern 1pt} {\kern 1pt} {\kern 1pt} {\kern 1pt} {\kern 1pt} {\kern 1pt} {\kern 1pt} {\kern 1pt} {\kern 1pt} {\kern 1pt} p \in {\bm{P}}_j^o}\\
				{\inf ,{\kern 1pt} {\kern 1pt} {\kern 1pt} {\kern 1pt} {\kern 1pt} {\kern 1pt} {\kern 1pt} {\kern 1pt} {\kern 1pt} {\kern 1pt} {\kern 1pt} {\kern 1pt} {\kern 1pt} {\kern 1pt} {\kern 1pt} {\kern 1pt} {\kern 1pt} {\kern 1pt} {\kern 1pt} {\kern 1pt} {\kern 1pt} {\kern 1pt} {\kern 1pt} {\kern 1pt} {\kern 1pt} {\kern 1pt} {\kern 1pt} {\kern 1pt} {\kern 1pt} {\kern 1pt} {\kern 1pt} {\bm{P}}_j^o{\rm{ = }}\phi }
		\end{array}} \right.
	\end{equation}

	And "Rangebin" $\bm{V_r}$ of each groundtruth is calculated by the same way.
	
	\begin{equation}
		\bm{V_r}[j] = num(\bm{P_j^g})
	\end{equation}	

	Where $\bm{P}_j^o$ and $\bm{P}_j^g$ are the obstacle points and the objects points in the $j$th column.
	
	\begin{figure}[t!]
		\centering
		\begin{tabular}{cc}
			\includegraphics[width=0.4\columnwidth]{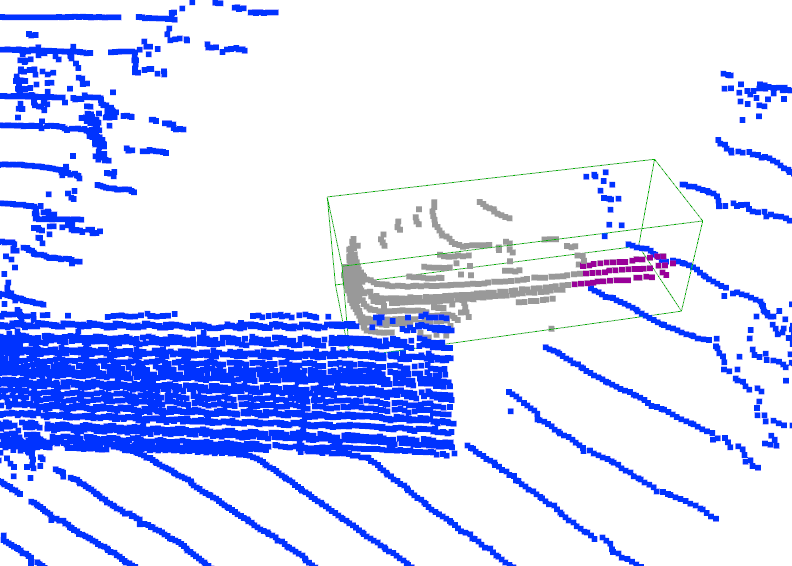}
			&
			\includegraphics[width=0.4\columnwidth]{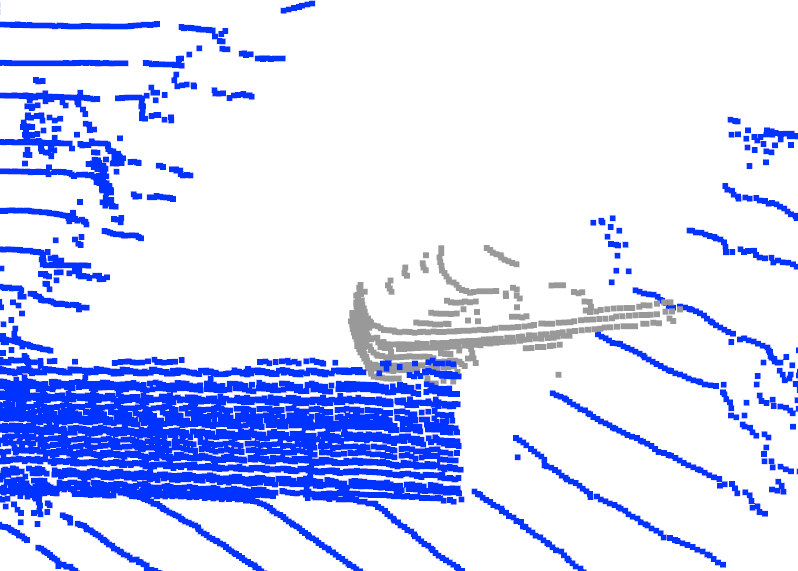}\\
			(a) Naive & (b) Culling\\
			\includegraphics[width=0.4\columnwidth]{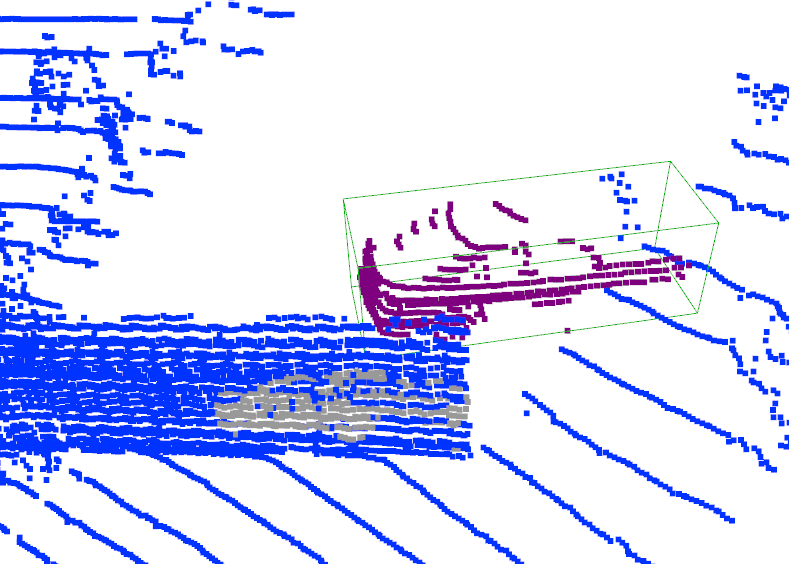}
			&
			\includegraphics[width=0.4\columnwidth]{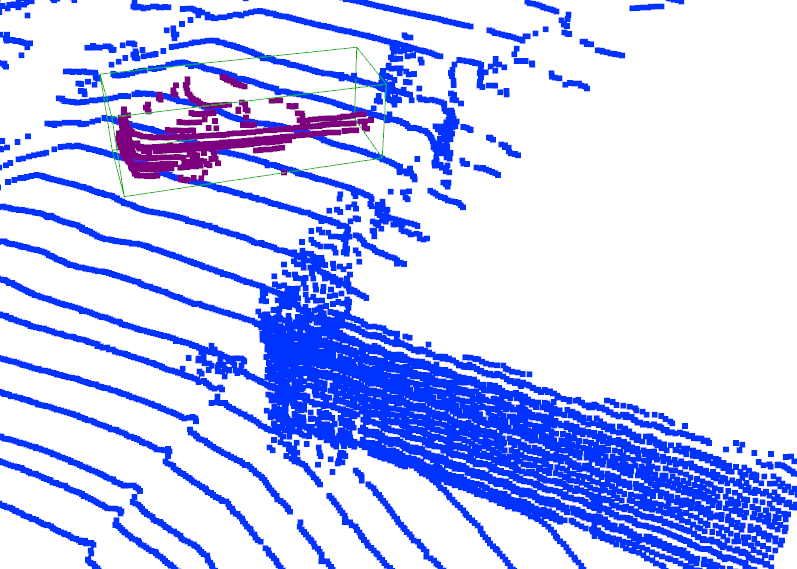}\\
			(c) Drilling & (d) CA-aug
		\end{tabular}
		\caption{Different types of augmentation for RV-based models. The gray points are deleted and the purple objects points are retained. (a) Directly project all points into the range image; (b) Discard the target if it losses too many points; (c) In the range image, the background points at the same pixels as the target points are deleted; (d) Move the target to another area to avoid excessive occlusion.}
		\label{fig:strategies}
	\end{figure}
	\setlength{\textfloatsep}{8pt}
	 
	\vspace{-1.5mm}
	\label{sec:CAD}

	\vspace{-0.2cm}
	\subsection{Object Placement}
	When obtaining "Validspace" and "Rangebin", we can quickly find reasonable locations by vector operation.
	
	Suppose that the start column of the inserted object is $j$ in the range image.We can estimate the rate of points in "Validspace":
	\begin{equation}
		\bm{M_j} = (\bm{V}[j:j + {l_g}] < {r_{box}} + {l_{box}}/2)\label{eq:judge1}
	\end{equation}
	\begin{equation}
		r_j = \bm{M_j}\cdot V_r/n_g\label{eq:judge2}
	\end{equation}

	Where $l_g$ is the length of $V_r$,$n_g$ is the number of object points and $l_{box}$ and $r_{box}$ are the length and the distance of the 3D box.
	The vector $M_j$ consists of 0 and 1, which indicates whether a column of object points are in "Validspace". When $r_j>a$, the angle corresponding to column $j$ is considered feasible.
	
	The detailed algorithm of location check is described in Algorithm \ref{alg:Location-search}. Note that simply adding the augmented objects into the background with the feasible angles will lead to a collision problem, which means that the 3D bounding boxes of the samples may intersect. Therefore, the collision avoidance algorithm need to be introduced. Suppose that there remain k bev boxes $b_1,...,b_k$ in the scene, when adding the next box $b_{k+1}$, we check the collision according to whether its four corners is inside the $k$ boxes. It's faster than calculating the IoU among all boxes.
	
	\begin{algorithm}[ht!]
		\caption{Location Check}
		\label{alg:Location-search}
		\textbf{Input:} -2D Freespace $\bm{V}$\\-Bev boxes array $\bm{B}$\\-Rangebin list $\bm{R}$;\\ 
		\textbf{Output:} -Boxes list $\bm{B_{out}}$; 
		\begin{algorithmic}[1]
			\STATE $N$ $\gets$ the number of boxes
			\STATE $i$ $\gets$ 0
			\STATE $W$ $\gets$ the length of $V$
			\WHILE {$i<N$}
			\STATE $l_g$ $\gets$ the length of $\bm{R}[i]$
			\STATE $n_g$ $\gets$ $sum(\bm{R}[i])$
			\STATE $\bm{A}$ $\gets$ $[\bm{V}[0:l_g],\bm{V}[1:l_g+1],...,\bm{V}[l_g:W-l_g]]$
			\STATE $\bm{M}$ $\gets$ $A<{r_{\bm{B}[i]}} + {l_{\bm{B}[i]}/2}$
			\STATE $\bm{r}$ $\gets$ $\bm{M}\cdot \bm{R}[i]/n_g$
			\STATE get all feasible start columns: $\bm{C}$ $\gets$ $where(\bm{r}>a)$
			\WHILE {$\bm{C}\neq \phi$}
			\STATE randomly select $c$ from $\bm{C}$ and rotate B[i] \\to $c$ around z axis 
			\IF{$\bm{B}[i]$ intersects any of $\bm{B}[0:i-1]$}
			\STATE delete $c$ in $\bm{C}$
			\ELSE
			\STATE break
			\ENDIF
			\ENDWHILE 
			\IF{$\bm{C}\neq \phi$}
			\STATE append $\bm{B}[i]$ in {$\bm{B_{out}}$}
			\ENDIF
			\ENDWHILE   
			\STATE
			\RETURN{$\bm{B_{out}}$}
		\end{algorithmic} 
	\end{algorithm}

	\section{Experiment}
	\begin{table*}[!t]
		\renewcommand\arraystretch{1.2}
		\centering
		\caption{The performance of CA-aug on the KITTI validation set. When training Rangetest, the "Culling" strategy is applied.}
		\label{tab:general}
		\setlength{\tabcolsep}{0.8mm}
		\resizebox{\textwidth}{!}{
			\begin{tabular}{l|l|ccc|ccc|ccc|c}
				\hline
				\multicolumn{1}{c|}{\multirow{2}{*}{Model}} & \multicolumn{1}{c|}{\multirow{2}{*}{Method}} & \multicolumn{3}{c|}{Car (IoU = 0.7)}  & \multicolumn{3}{c|}{Pedestrian (IoU = 0.5)} & \multicolumn{3}{c|}{Cyclist (IoU = 0.5)} & \multicolumn{1}{c}{mAP}     
				\\ 
				\multicolumn{1}{c|}{}  & \multicolumn{1}{c|}{} & \multicolumn{1}{c}{Easy} & \multicolumn{1}{c}{Moderate} & \multicolumn{1}{c|}{Hard} & \multicolumn{1}{c}{Easy} & \multicolumn{1}{c}{Moderate} & \multicolumn{1}{c|}{Hard} & \multicolumn{1}{c}{Easy} & \multicolumn{1}{c}{Moderate} & \multicolumn{1}{c|}{Hard} & \multicolumn{1}{c}{Moderate} \\ 
				\hline
				\multicolumn{1}{c|}{\multirow{2}{*}{PointPillar}} & Baseline & \textbf{87.76} & \textbf{78.26} & 75.34 & 55.49 & 50.27 & 45.63 & 79.21 & 60.93 & 57.35 & 63.15 \\ 
				~ & CA-aug & 87.61(-0.15) & \textbf{78.26}(+0.00) & \textbf{75.45}(+0.09) &\textbf{58.32}(+2.83) & \textbf{52.49}(+2.22) & \textbf{48.24}(+2.61) & \textbf{81.81}(+2.60) & \textbf{62.66}(+1.73) & \textbf{58.89}(+1.54) & \textbf{64.47}(+1.32) \\ 
				\hline
				\multicolumn{1}{c|}{\multirow{2}{*}{SECOND}} & Baseline & \textbf{90.77} & \textbf{81.95} & \textbf{78.91} & 56.88 & 52.96 & 48.22 & 82.40 & 64.09 & 59.69 & 66.33 \\ 
				~ & CA-aug & 90.55(-0.22) & 81.47(-0.48) & 78.60(-0.31) & \textbf{58.91}(+2.03) & \textbf{53.93}(+0.97) & \textbf{49.29}(+1.07) & \textbf{87.85}(+5.45) & \textbf{69.75}(+5.66) & \textbf{65.46}(+5.77) & \textbf{68.38}(+2.05) \\ 
				\hline
				\multicolumn{1}{c|}{\multirow{2}{*}{PVRCNN}} & Baseline & 91.76 & 84.62 & 82.45 & 63.63 & 56.81 & 52.14 & 88.78 & 70.84 & 66.58 & 70.75 \\ 
				~ & CA-aug & \textbf{92.25}(+0.49) &\textbf{84.93}(+0.31) & \textbf{82.64}(+0.19) & \textbf{66.00}(+2.37) & \textbf{59.77}(+2.96) & \textbf{55.41}(+3.28) & \textbf{92.58}(+3.80) & \textbf{74.19}(+3.35) & \textbf{69.76}(+3.18) & \textbf{72.96}(+2.21) \\ 
				\hline
				\multicolumn{1}{c|}{\multirow{2}{*}{Rangetest}} & Baseline & 88.66 & 77.44 & 72.74 & 47.76 & 39.55 & 35.11 & 71.25 & 50.83 & 46.38 & 55.94 \\ 
				~ & CA-aug & \textbf{89.99}(+1.33) & \textbf{80.52}(+3.08) & \textbf{75.64}(+3.10) & \textbf{52.91}(+5.15) & \textbf{42.22}(+2.67) & \textbf{35.63}(+0.52) & \textbf{77.49}(+6.24) & \textbf{58.66}(+7.83) & \textbf{54.03}(+7.65) & \textbf{60.46}(+4.52) \\ \hline
			\end{tabular}
		}
	\vspace{-1.0em}
	\end{table*}
	\subsection{Experiment Setup}
	\noindent
	{\bf Dataset}\hspace{0.3cm} We trained and evaluated all models on KITTI, one of the most popular datasets of 3D object detection. It contains 7481 training frames and 7518 testing frames. By convention, the training frames are further divided into the training split(3712 samples) and the validation split(3769 samples).We test our method on val split. There are three categories of targets: cars, pedestrians, and cyclists. For each, three difficulty levels (easy, moderate, hard) are involved. We apply KITTI's evaluation benchmark - 3D average precision(AP) calculated with 40 recall positions to evaluate the detection results.
	
	\noindent 
	{\bf Models}\hspace{0.3cm} 
	To demonstrate the universality of CA-aug, we conduct experiments on four models. For the RV-based model, we remove the RCNN module of and name this one-stage detector Rangetest. All the models are implemented with default parameters from their original papers\cite{Lang2019PointPillarsFE,liangRangeRCNNFastAccurate2021,Shi2020PVRCNNPF,yanSECONDSparselyEmbedded2018}.
	
	\noindent
	{\bf Augmentation}\hspace{0.3cm} Besides GT-sample, we compare CA-aug with "Validmap" and the three Augmentation strateies for RV-models. For fairness, No CAD models are introduced to extend groundtruth database. For "Validmap", we applied the original hyparameters in 3. For "Culling", we remove objects that remain less than 4 or lose 75\% surface points. For CA-aug, the pillar length $d$ is set to $0.25$m and $\sigma=0.4$m. We set the threshold $a=0.8$ to allow objects appearing behind thin obstacles like rods. All the models are also trained with global augmentations including random rotation, flipping and scaling.
		
	\subsection{Comparison}
	In Table \ref{tab:comparsion1} and Table \ref{tab:comarsion_range}, we report the results of comparative experiments. The "baseline" method means that no objects are inserted. As can be observed, CA-aug achieves the best performance in Cars detection accuracy. 
	
	Although the "Validmap" ensures the rationality of the objects' positions, their ranges are randomly selected and the geometric characteristics of scanning lines are not maintained. Putting dense targets in the distance or sparse ones in near places could mislead the network into learning the information that the density of foreground points is independent of their distances. However, "Validmap" is suitable for the placement of CAD models rather than groundtruths, because appropriate point patterns can be generated on the surface of CAD models according to their ranges. CA-aug takes advantage of the radial symmetry of lidar points and changes the sample positions through rotation. So the above issue does not occur.
	\begin{table}[h!]
		\setlength{\tabcolsep}{4pt}
		\vspace{-0.3em}
		\begin{center}
			\caption{The comparsion of CA-aug and Validmap. The 3D AP performance of PointPillar is reported.}
			\begin{tabular}{l|ccc}   
				\hline
				\multicolumn{1}{c|}{\multirow{2}{*}{Method}} & \multicolumn{3}{c}{Car (IoU = 0.7)}  \\
				
				\multicolumn{1}{c|}{}   & \multicolumn{1}{c}{Easy} & \multicolumn{1}{c}{Moderate} & \multicolumn{1}{c}{Hard} \\ 
				\hline
				Baseline & 88.08 & 74.85 & 70.55  \\\hline
				GT-sample & 87.80 & 78.36 & 75.41 \\\hline
				Validmap & 87.75 & 78.24 & 75.35 \\\hline
				CA-aug(ours) & \textbf{88.82} & \textbf{78.66} & \textbf{75.75} \\
				\hline
			\end{tabular}   
			\label{tab:comparsion1}
		\end{center}
	\end{table}
	
	\begin{table}[h!]
		\setlength{\tabcolsep}{5pt}
		\vspace{-3.0em}
		\begin{center}
			\caption{Comparsion of CA-aug and the other three strategies. The 3D AP performance of Rangetest is reported.}
			\begin{tabular}{l|ccc}   
				\hline
				\multicolumn{1}{c|}{\multirow{2}{*}{Method}} & \multicolumn{3}{c}{Car (IoU = 0.7)}\\ 
				\multicolumn{1}{c|}{} & \multicolumn{1}{c}{Easy} & \multicolumn{1}{c}{Moderate} & \multicolumn{1}{c}{Hard} \\ 
				\hline
				Baseline & 85.14 & 73.45 & 68.64 \\  \hline
				Naive & 88.47 & 77.69 & 72.92  \\\hline
				Culling & 89.15 & 78.28 & 73.32 \\ \hline
				Drilling & 88.84 & 78.17 & 73.27 \\ \hline
				CA-aug(ours) & \textbf{90.33} & \textbf{81.04} & \textbf{76.06} \\\hline
			\end{tabular}   
			\label{tab:comarsion_range}
			\vspace{-3.0em}
		\end{center}
	\end{table}
	\begin{table}[h!]
		\setlength{\tabcolsep}{5pt}
		\begin{center}
			\caption{The further experiment on controling occlusion. The 3D AP performance of Rangetest is reported.}
			\begin{tabular}{l|ccc}   
				\hline
				\multicolumn{1}{c|}{\multirow{2}{*}{Method}} & \multicolumn{3}{c}{Car (IoU = 0.7)}\\ 
				\multicolumn{1}{c|}{} & \multicolumn{1}{c}{Easy} & \multicolumn{1}{c}{Moderate} & \multicolumn{1}{c}{Hard} \\ 
				\hline
				CA-aug & \textbf{90.33} & \textbf{81.04} & \textbf{76.06} \\\hline
				CA-aug + Drilling & 89.93 & 80.76 & 75.80  \\\hline
				CA-aug + Update + Drilling& 90.05 & 80.58 & 73.60 \\  \hline
			\end{tabular}   
			\label{tab:further}
			\vspace{-1.0em}
		\end{center}
	\end{table}
	Compared to Naive, both Culling and Drilling alleviate the loss of surface points, enabling the network to learn more information from relatively complete targets. By moving objects to reasonable places, CA-aug not only controls the degree of occlusion, but also guides the model to explore more semantic information between objects and backgrounds. 
	
	After moving to "Validspace", objects could still be occluded by closer points of the environment or other targets, which has uncertain effects. On the one hand, it may force the network to improve the ability to infer the complete shapes of targets from sparse points. On the other hand, incomplete targets may hurt the training process. It motivates a further experiment whose result is reported in \ref{tab:further}. We add Drilling to avoid occlusion by background. Meanwhile, To control the occlusion by other targets, during the location search, we place the bev target boxes in the order of their center distance from near to far and update the "Validspace" using the following equation. 
	\begin{equation}
		\bm{V}[j:j + {l_g}] = r_{box}
	\end{equation}

	After adding Drilling, the AP for all difficulty levels decreases. "Update" even reduces the AP for hard targets by 2.2. This indicates the positive effects of proper occlusion.

	\subsection{Generalization}
	As shown in Table \ref{tab:general}, our method achieves higher mAP for all models. For the detection of cyclists and pedestrians, it outperforms GT-sample  
	by a large margin mainly because it can mitigate the second side effect of GT-sample mentioned in Section \ref{sec:intro}. However, CA-aug doesn't provide SECOND and PointPillar with better performance of car detection, since it makes them pay more attention to the other two classes. For heavy model PV-RCNN, CA-aug brings an overall AP improvement, which indicates its ability to reduce overfitting. CA-aug can also exploits the potential of the RV-based model, increasing Rangetest' s mAP by nearly 8 \%.
	\begin{table}[h!]
		\setlength{\tabcolsep}{2.5pt}
		\vspace{-1.0em}
		\begin{center}
			\caption{The ablation study on CA-aug using Rangetest model}
			\begin{tabular}{cccc|ccc}   
				\hline
				\multicolumn{1}{c}{\multirow{2}{*}{Baseline}} &
				\multicolumn{1}{c}{\multirow{2}{*}{Collision Avoid}} &
				\multicolumn{1}{c}{\multirow{2}{*}{Rotate}} & \multicolumn{1}{c|}{\multirow{2}{*}{Culling} }& \multicolumn{3}{c}{Car (IoU = 0.7)}\\ 
				\multicolumn{1}{c}{}&\multicolumn{1}{c}{}&\multicolumn{1}{c}{}&\multicolumn{1}{c|}{}& \multicolumn{1}{c}{Easy} & \multicolumn{1}{c}{Moderate} & \multicolumn{1}{c}{Hard} \\ 
				\hline
				\checkmark & \multicolumn{1}{c}{-}&\multicolumn{1}{c}{-} & \multicolumn{1}{c|}{-} & 88.29 & 77.67 & 72.89 \\\hline
				\checkmark & \checkmark&\multicolumn{1}{c}{-} & \multicolumn{1}{c|}{-} & 89.79 & 78.36 & 73.37 \\\hline
				\checkmark & \checkmark&\checkmark &\multicolumn{1}{c|}{-} & 89.44 & 80.14 & 75.24  \\\hline
				\checkmark &\checkmark&\checkmark &\checkmark& \textbf{90.33} & \textbf{81.04} & \textbf{76.06}   \\\hline
			\end{tabular}   
			\label{tab:Ablation}
			\vspace{-3.0em}
		\end{center}
	\end{table}
	\subsection{Ablation Study}
	
	The result of the ablation study is shown in Table \ref{tab:Ablation}. If the "Rotate" component is not chosen, we just remove the objects whose original places are judged to be inreasonable by Equation \ref{eq:judge1},\ref{eq:judge2}.
	Without enforcing collision avoidance, the overall AP decreases. The "Rotate" operation improves the moderate and hard AP by 1.78 and 1.87, while incorporating \textit{Culling} can further boost the AP by 0.89, 0.9 and 0.82.
	
	\section{Conclusion}
	\vspace{-0.2cm}
	\label{sec:con}
	We have represented CA-aug to solve the problem of the irrational target placement in GT-sample. It can create more realistic augmented 3D scenes to help models exlore the semantic relationships between targets and environments. Our method is lightweight and compatible with other augmentations. the experiments demonstrate its effectiveness and generalization. 
	
	Like \cite{fangLiDARAugGeneralRenderingbased2021,manivasagamLiDARsimRealisticLiDAR2020}, CA-aug can also be used to determine the places of synthetic objects. We believe that CA-aug is applicable to other tasks in the field of autonomous driving, such as 3D object tracking and 3D semantic segmentation and we will do further research on it.
	
	
\bibliographystyle{plain}
\bibliography{ca.bib}
\end{document}